\documentclass[11pt,a4paper]{article}
\usepackage[hyperref]{emnlp2018}
\usepackage{times}
\usepackage{latexsym}

\usepackage{graphicx}
\usepackage{wrapfig}
\usepackage{url}
\usepackage{amsmath}
\usepackage{amssymb}
\usepackage{color,xcolor,colortbl}
\usepackage{enumitem}
\usepackage{algorithm}
\usepackage{algorithmic}
\usepackage[font={small}]{caption}
\usepackage{bm,bbm}
\usepackage{booktabs}
\usepackage{mathtools}
\usepackage{array}

\newcommand\numberthis{\addtocounter{equation}{1}\tag{\theequation}}

\definecolor{darkblue}{rgb}{0.0, 0.0, 0.55}
\newenvironment{fontppl}{\fontfamily{ppl}\selectfont}{\par} 

\aclfinalcopy 
\def\aclpaperid{1340} 

\title{Automatic Detection of Vague Words and Sentences in Privacy Policies}

\author{Logan Lebanoff \and Fei Liu\\ 
  Department of Computer Science\\
  University of Central Florida, 
  Orlando, FL 32816, USA\\
  {\tt loganlebanoff@knights.ucf.edu \quad feiliu@cs.ucf.edu}}

\date{}

\begin{document}

\maketitle

\begin{abstract}

Website privacy policies represent the single most important source of information for users to gauge how their personal data are collected, used and shared by companies. 
However, privacy policies are often vague and people struggle to understand the content. 
Their opaqueness poses a significant challenge to both users and policy regulators. 
In this paper, we seek to identify vague content in privacy policies.
We construct the first corpus of human-annotated vague words and sentences and present empirical studies on automatic vagueness detection.
In particular, we investigate context-aware and context-agnostic models for predicting vague words, and explore auxiliary-classifier generative adversarial networks for characterizing sentence vagueness.
Our experimental results demonstrate the effectiveness of proposed approaches.
Finally, we provide suggestions for resolving vagueness and improving the usability of privacy policies.

\end{abstract}

\section{Introduction}

Website privacy policies are difficult to read and people struggle to understand the content.
Recent studies~\cite{Sadeh:2013} have raised concerns over their opaqueness, which poses a considerable challenge to both Internet users and policy regulators.
Nowadays, consumers supply their personal information to online websites in exchange for personalized services;
they are surrounded by smart gadgets such as voice assistants and surveillance cameras, which constantly monitor their activities in the home and work environments.
Without clearly specifying how users' information will be collected, used and shared, there is a substantial risk of information misuse, including undesired advertisements and privacy breaches.
Especially with recent high-profile cases involving Facebook and Cambridge Analytica, the public is becoming more aware and concerned with how their information is handled.

Privacy policies are binding agreements between companies and users that stipulate how companies collect, use, and share users' personal information. 
They are lengthy and difficult to read.
Bhatia et al.~\shortcite{Bhatia:2016} suggested two possible causes for this.
First, privacy policies must be \emph{comprehensive} in order to cover a variety of uses (e.g., in-store and online purchases).
Second, the policies have to be \emph{accurate} to all data practices and systems. 
Clearly, it would be difficult for a company's legal counsel to anticipate all future needs.
They need to resort to vague language to describe the content, causing it to be difficult to read and compromising the effectiveness of privacy policies.

\begin{table}[t]
\setlength{\tabcolsep}{5pt}
\renewcommand{\arraystretch}{1.1}
\centering
\begin{footnotesize}
\begin{fontppl}
\begin{tabular}{|l|p{2.5in}|}
\hline
S1 & We \textcolor{red}{\textbf{\emph{may}}} use the \textcolor{red}{\textbf{\emph{information}}} automatically collected from your computer or \textcolor{red}{\textbf{\emph{other devices}}} for the following uses... (Vagueness: 3.8)\\
\hline
\hline
S2 & In addition, in \textcolor{red}{\textbf{\emph{some}}} cases the Sites can deliver content based on your current location if you choose to enable that feature.\\ & (Vagueness: 2.25)\\
\hline
\hline
S3 & Our Sites and Services \textcolor{red}{\textbf{\emph{may}}}, \textcolor{red}{\textbf{\emph{from time to time}}}, provide links to sites operated by \textcolor{red}{\textbf{\emph{third parties}}}. (Vagueness: 3.2)\\
\hline
\hline
S4 & To customize and serve advertising and \textcolor{red}{\textbf{\emph{other}}} marketing communications that \textcolor{red}{\textbf{\emph{may}}} be visible to you on our Sites and Services or \textcolor{red}{\textbf{\emph{elsewhere}}} on the internet. (Vagueness: 4)\\
\hline
\hline
S5 & This includes your credit card number, income level, or \textcolor{red}{\textbf{\emph{any other}}} information that would \textcolor{red}{\textbf{\emph{normally}}} be considered confidential.\\ & (Vagueness: 3)\\

\hline
\end{tabular}
\end{fontppl}
\end{footnotesize}
\caption{Example human-annotated vague words and sentences. Vague words are \emph{italicized}. Averaged sentence vagueness is given in the parentheses. Higher score is more vague.
}
\label{tab:example}
\vspace{-0.2in}
\end{table}

In this paper, we present the first study on automatic detection of vague content in website privacy policies. 
We construct a sizable corpus containing word- and sentence-level human annotations of vagueness for privacy policy documents. 
The corpus contains a total of 133K words and 4.5K sentences. 
Our methods for automatically detecting vague words and sentences are based on deep neural networks, which have demonstrated impressive recent success.
Specifically, we investigate context-aware and context-agnostic models for predicting word vagueness, where feature representations of words are built with and without considering their surrounding words.
By this, we seek to verify the hypothesis that vagueness is an intrinsic property of words and has little to do with context.
To understand sentence vagueness, we explore auxiliary-classifier generative adversarial networks (AC-GAN, Odena et al., 2018\nocite{Odena:2018}).
The model has performed strongly on vision tasks (e.g., image synthesis), however, whether it can be adapted to handle text data has not been thoroughly investigated. 
We train the AC-GAN model to discriminate between real/fake privacy policy sentences while simultaneously classifying sentences exhibiting different levels of vagueness, including ``clear,'' ``somewhat clear,'' ``vague,'' and ``extremely vague,'' thus improving the model's generalization capabilities.
The detected vague words and sentences can assist users in browsing privacy policy documents, and privacy regulators in assessing the clarity of privacy policy practices. 
Our research contributions include the following:
\begin{itemize}[topsep=3pt,itemsep=-1pt,leftmargin=*]

\item we present the first study on automatic detection of vague content in privacy policies. 
Vague content compromises the usability of privacy policies and there is an urgent need to identify and resolve vagueness;

\item we construct a sizable text corpus including human annotations for 133K words and 4.5K sentences of privacy policy texts. 
The data\footnote{\url{https://loganlebanoff.github.io/data/vagueness_data.tar.gz}} is available publicly to advance research on language vagueness; and

\item we investigate both context-aware and context-agnostic methods for predicting vague words. We also explore the auxiliary-classifier generative adversarial networks for characterizing sentence vagueness. This is the first study leveraging deep neural networks for detecting vague content in privacy policies. 

\end{itemize}

\section{Related Work}

Privacy policies are often verbose, difficult to read, and perceived as ineffective \cite{Mcdonald:2008}.
In particular, vague language in these documents hurts understanding. \emph{``A term is regarded as vague if it admits borderline cases, where speakers are reluctant to say either the term definitely applies or definitely does not apply,''} a definition of vagueness quoted from~\cite{Deemter:2010}.
Legal scholars and language philosophers strive to understand vagueness from a theoretical perspective~\cite{Keefe:2000,Shapiro:2006}.
The ``sorites paradox'' describes the phenomenon of vagueness~\cite{Keefe:2000}.
It states that small changes in the object do not affect the applicability of a vague term.
For example, a room can remain ``bright'' even if the light is dimmed little by little until it is entirely extinguished, thus creating a paradox. 
Hyde~\shortcite{Hyde:2014} further suggests that vagueness is a feature pertaining to multiple syntactic categories. 
Nouns, adjectives and adverbs (e.g., ``child'', ``tall'', ``many'') are all susceptible to reasoning.
These studies often focus on linguistic case studies but not on developing resources for automatic detection of vagueness.

Recent years have seen a growing interest in using natural language processing techniques to improve the effectiveness of website privacy policies.
Sadeh et al.~\shortcite{Sadeh:2013} describe a Usable Privacy Policy Project that seeks to semi-automate the extraction of salient details from privacy policies.
Other studies include crowdsourcing privacy policy annotations and categorizing data practices~\cite{Ammar:2012,Massey:2013,Wilson:2016:WWW,Wilson:2016:ACL}, grouping text segments related to certain policy issues~\cite{Liu:2014:COLING,Ramanath:2014:ACL}, summarizing terms of services~\cite{Braun:2017}, identifying user opt-out choices~\cite{Sathyendra:2017}, and many others.
These studies emphasize the ``too long to read'' issue of privacy policies but leave behind the ``difficult to understand'' aspect, such as identifying and eliminating vague content. 

The work of~\cite{Liu:2016:PLT} is close to ours.
The authors attempt to learn vector representations of words in privacy policies using deep neural networks, where the vectors encode not only semantic/syntactic aspects but also vagueness of words.
The model is later fed to an interactive visualization tool~\cite{Strobelt:2016} to test its ability to discover related vague terms.
While promising, their approach is not fully automatic, and the feasibility of detecting vague words and sentences in an automatic manner is still left untested.

In this work we conduct the first study to automatically detect vague content from privacy policies.
We ask human annotators to label vague words and sentences and train supervised classifiers to do the same.
Classifying vague words is a challenging task, because vagueness is an understudied  property and it spans multiple syntactic categories (e.g., ``usually,'' ``personal data,'' ``necessary'').
Neural network classifiers such as CNN and LSTM have demonstrated prior success on text classification tasks~\cite{Zhang:2015}, but whether they can be utilized to identify vague terms is not well understood.

For sentence classification, we investigate auxiliary classifier generative adversarial networks (AC-GAN, Odena et al., 2018\nocite{Odena:2018}).
GANs have seen growing popularity in recent years~\cite{Mirza:2014,Yu:2016,Li:2017:GAN,Gu:2018,Cai:2018}.
AC-GAN is a variant of GAN that generates word sequences using class-conditional probabilities. 
E.g., it generates ``fake'' privacy policy sentences exhibiting different degrees of vagueness (e.g., ``clear,'' ``vague,'' ``extremely vague'').
AC-GAN nicely combines real (human-annotated) and fake (synthetic) privacy policy sentences in a discriminative framework to improve the model's generalization capabilities.
This can be equated to a semi-supervised learning paradigm through augmentation of the dataset with generated sentences.
Data augmentation is particularly valuable for vagueness detection, which generally has small expensive datasets.
We perform a full analysis on AC-GAN and compare it to state-of-the-art systems.

\section{The Corpus}

\begin{table}[t]
\setlength{\tabcolsep}{3pt}
\renewcommand{\arraystretch}{1.1}
\centering
\begin{footnotesize}
\begin{fontppl}
\begin{tabular}{lr|lr}
\bottomrule[0.3mm]
\textbf{Vague Term} & \textbf{Freq.} & \textbf{Vague Term} & \textbf{Freq.}\\
\hline
may & 1,575 & other information & 30\\
personal information & 465 & non-personal info. & 30\\
information & 302 & sometimes & 27 \\
other & 261 & reasonably & 26\\
some & 214 & appropriate & 25\\
certain & 205 & necessary & 24\\
third parties & 183 & certain information & 23\\
third party & 134 & typically & 22\\
personally iden. info. & 88 & affiliates & 21\\
time to time & 75 & reasonable & 20\\
most & 54 & non-personal & 19\\
generally & 52 & personally iden. & 18\\
personal data & 52 & such as & 18\\
third-party & 49 & usually & 17\\
others & 41 & personal & 16\\
general & 39 & may be & 15\\
many & 37 & content & 14\\
various & 36 & otherwise & 14\\
might & 35 & periodically & 14\\
services & 33 & similar & 14\\
\toprule[0.3mm]
\end{tabular}
\end{fontppl}
\end{footnotesize}
\caption{The most frequent vague terms identified by human annotators and their frequencies in our corpus. ``iden.'' and ``info.'' are shorthand for ``identifiable'' and ``information.''
}
\label{tab:vague_terms}
\vspace{-0.15in}
\end{table}

Annotating vague words and sentences is a nontrivial task.
We describe our effort to select privacy policy sentences for annotation, recruit qualified workers, and design annotation guidelines.

\begin{figure}
\centering
\includegraphics[width=3.1in]{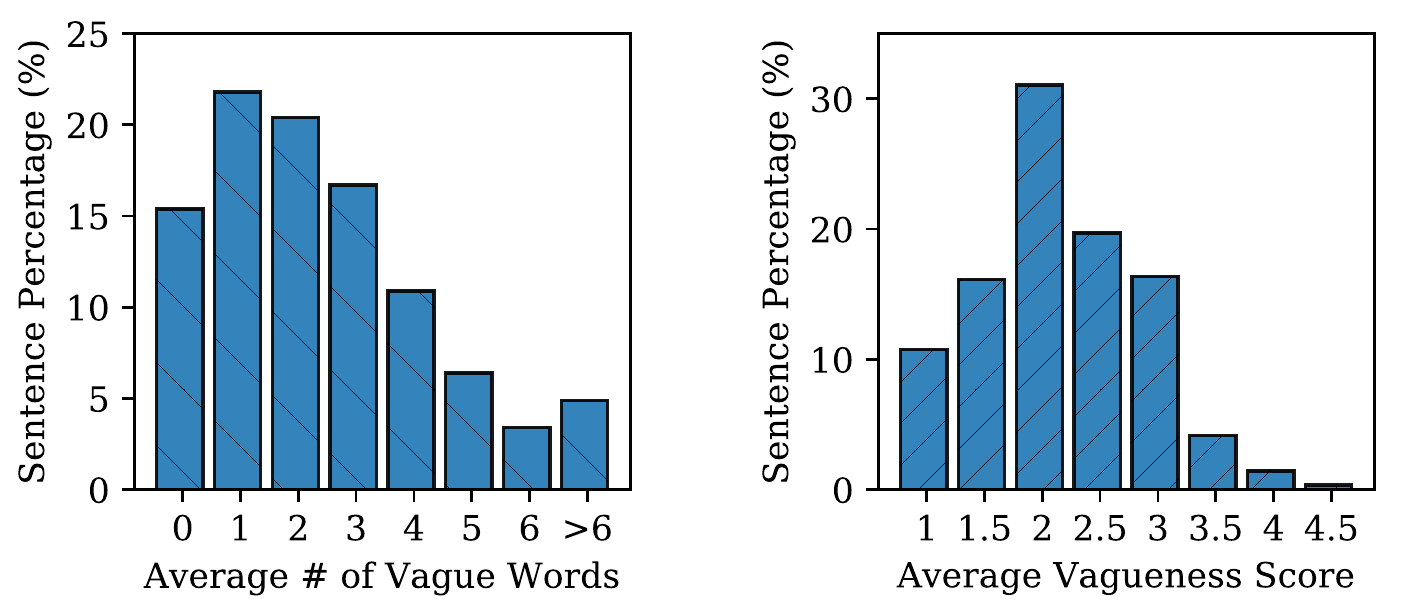}
\caption{(Left) Percentage of sentences containing different numbers of vague words. (Right) Perc. of sentences with different levels of vagueness. 1 is clear, 5 is extremely vague.}
\label{fig:plot_vague}
\vspace{-0.1in}
\end{figure}

We select 100 website privacy policies from the collection gathered by Liu et al.~\shortcite{Liu:2014:COLING}.
The documents are quite lengthy, containing on average 2.3K words.
More importantly, most content is not vague. 
To obtain a more balanced corpus, a filtering step is used to select only sentences that have a moderate-to-high chance of containing vague content.
Fortunately, Bhatia et al.~\shortcite{Bhatia:2016} provide a list of 40 cue words for vagueness, manually compiled by policy experts.
We therefore retain only sentences containing one of the cue words for further annotation.
A brief examination shows that most of the sentences removed from the corpus are indeed clear.
Even with this bias, the resulting corpus still contains a small portion of clear sentences (See Figure~\ref{fig:plot_vague}).
The reason is that a cue word can be used in a way that is not vague.
For example, in the sentence ``Users \textit{may} post to our website,'' the word \emph{may} indicates permission but not possibility, and therefore the sentence is not vague.

Reidenberg et al.~\shortcite{Reidenberg:2015} discuss attempts to use crowdsourced workers as a cost-effective alternative to policy experts for annotating privacy policies.
In this study, we hire crowd workers from the Amazon Mechanical Turk platform. 
To recruit quality workers, we require them to reside in the U.S. and be proficient in English; they are skilled workers maintaining a task success rate of 90\% or above. 
We provide example labelled vague terms obtained from the case studies described in Bhatia et al.~\shortcite{Bhatia:2016} to reduce discrepancies among workers. 
The annotators are then asked to use their best judgment to perform the task.

Given a privacy policy sentence, the annotators are instructed to identify all vague terms\footnote{We use ``term'' to denote either a single word or a phrase.} and assign a score of vagueness to the sentence.
A vague term is limited to be 5 words or less (e.g., ``including but not limited to'').
We use this rule to prevent annotators from tagging an entire sentence/clause as vague.
A slider is provided in the interface to allow annotators to select a vagueness score for the sentence: 1 is extremely clear and 5 is extremely vague.
We design a human intelligence task (HIT) to include 5 privacy policy sentences and a worker is rewarded \$0.05 for completing the task. 
Five human workers are recruited to perform each task.

We obtain annotations for 133K words and 4.5K sentences.
The average sentence vagueness score is 2.4$\pm$0.9.
As of inter-annotator agreement, we find that
47.2\% of the sentences have their vagueness scores agreed by 3 or more annotators;
12.5\% of the sentence vagueness scores are agreed by 4 or more annotators.
Furthermore, the annotators are not required to select vague words if they believe the sentences are clear. 
We remove vague words selected by a single annotator. Among the rest, 46.1\% of the words are selected by 3 or more annotators;
18.5\% of the words are selected by 4 or more annotators.
These results suggest that, although annotating vague terms and sentences is considered challenging, our annotators can reach a reasonable degree of agreement.\footnote{We choose not to calculate a kappa statistic, because labelling vague words/sentences is not a clear-cut classification task and it is difficult to apply kappa to this setting.}
We present example vague terms in Table~\ref{tab:vague_terms}.
Note that we obtain a total of 1,124 unique vague terms, which go well beyond the 40 cue words used for sentence preselection.
Figure~\ref{fig:plot_vague} shows more statistics on sentence vagueness, including 
(i) the percentages of sentences containing different numbers of vague words, and (ii) the percentages of sentences whose vagueness scores fall in different ranges.

\section{Word Vagueness}

We seek to test an important hypothesis related to word vagueness. 
We conjecture that vagueness is an intrinsic property of words; whether a word is vague or not has little to do with its context words.
To verify this hypothesis, we build context-aware and context-agnostic models to classify each word in a privacy policy sentence as either vague or non-vague. 
The ground-truth labels are obtained by consolidating human annotations (see Table~\ref{tab:word_labels} for an example). 
A word is labelled 1 if it is selected by two or more annotators, otherwise 0.
We describe details of the two classifiers below.

\begin{figure}
\centering
\includegraphics[width=2.5in]{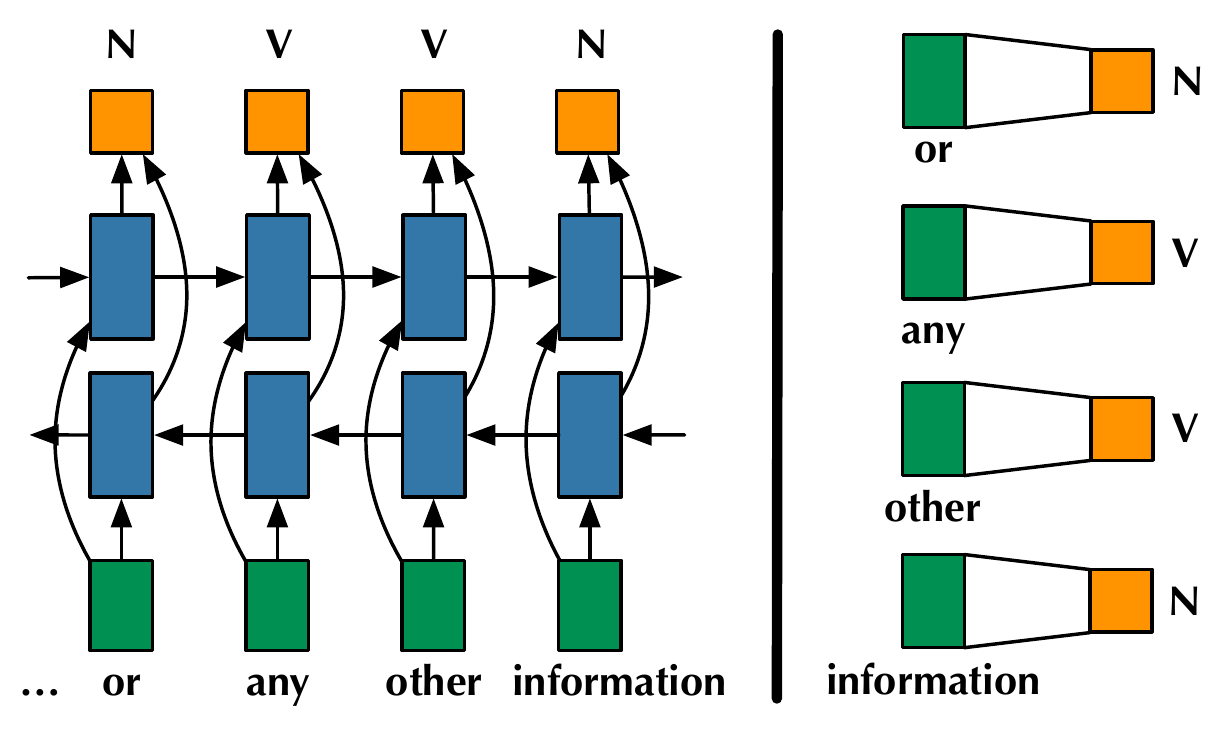}
\caption{(Left) Context-aware word classifier implemented as a bidirectional LSTM. (Right) Context-agnostic classifier.  ``V'' and ``N'' are shorthands for ``vague'' and ``non-vague.''
}
\label{fig:word_classifiers}
\vspace{-0.2in}
\end{figure}

\vspace{0.05in}
\noindent\textbf{Context-aware classifier.}
It builds feature representations of words based on the surrounding context words.
Given its strong performance, we construct a bidirectional LSTM~\cite{Hochreiter:1997} for this purpose.
A word is replaced by its word2vec embedding~\cite{Mikolov:2013} before it is fed to the model.
For each time step, we concatenate the hidden states obtained from the forward and backward passes and use it as input to a feedforward layer with sigmoid activation to predict if a word is vague or non-vague.
Because single words consist of the majority of the human-annotated vague terms, we choose to use binary word labels instead of a BIO scheme~\cite{Chiu:2016} for sequence tagging.
Figure~\ref{fig:word_classifiers} shows the architecture.

\vspace{0.05in}
\noindent\textbf{Context-agnostic classifier.}
It uses intrinsic feature representations of words without considering the context.
Specifically, we represent a word using its word2vec embedding, then feed it to a feedforward layer with sigmoid activation to obtain the prediction (Figure~\ref{fig:word_classifiers}).
We train the classifier using a list of unique words obtained from the training data; a word is considered positive if it has a ground truth label of 1 in any sentence, otherwise negative. 
Note that the ratio of positive/negative unique words in our corpus is 1068/3176=0.34. 
At test time, we apply the binary classifier to each word of the test set.
A word is assigned the same label regardless of which sentence it appears in.
We adopt this setting to ensure the context-aware and context-agnostic results are comparable.

\begin{table}[t]
\setlength{\tabcolsep}{5pt}
\renewcommand{\arraystretch}{1.1}
\centering
\begin{scriptsize}
\begin{fontppl}
\begin{tabular}{|p{2.9in}|}
\hline
\textbf{Sent:} This includes your credit card number , income level , or any other information that would normally be considered confidential .\\ 
\hline
\hline
\textbf{Annotator 1:} any, other, normally\\
\textbf{Annotator 2:} any other information\\
\textbf{Annotator 3:} normally, confidential, any other\\
\hline
\hline
\textbf{Ground Truth Labels:} [This]$_0$ [includes]$_0$ [your]$_0$ [credit]$_0$ [card]$_0$ [number]$_0$ , [income]$_0$ [level]$_0$ , [or]$_0$ \textcolor{black}{\textbf{\emph{[any]}}}$_{1}$ \textcolor{black}{\textbf{\emph{[other]}}}$_{1}$ [information]$_{0}$ [that]$_{0}$ [would]$_{0}$ \textcolor{black}{\textbf{\emph{[normally]}}}$_{1}$ [be]$_{0}$ [considered]$_{0}$ [confidential]$_{0}$ .\\ 
\hline
\end{tabular}
\end{fontppl}
\end{scriptsize}
\caption{Ground truth labels are obtained by consolidating human-annotated vague terms; ``any,'' ``other,'' ``normally'' are labelled 1 because they are selected by 2 or more annotators.
}
\label{tab:word_labels}
\vspace{-0.1in}
\end{table}

\vspace{-0.05in}
\section{Sentence Vagueness}

We next investigate how vagueness is manifested in privacy policy sentences.
Our goal is to assign a label to each sentence indicating its level of vagueness.
We derive ground truth sentence labels by averaging over vagueness scores assigned by human annotators, and further discretizing the scores into four buckets: [1,2), [2,3), [3,4), [4,5], respectively corresponding to ``clear,'' ``somewhat clear,'' ``vague,'' and ``extremely vague'' categories. 
The sentences in the four buckets respectively consist of 26.9\%, 50.8\%, 20.5\%, and 1.8\% of the total annotated sentences. 
We choose to predict discrete labels instead of continuous scores because labels are more informative to human readers. E.g., a label of ``extremely vague'' is more likely to trigger user alerts than a score of 4.2.

\subsection{Auxiliary-Classifer GAN}

\begin{figure}
\centering
\includegraphics[width=2.9in]{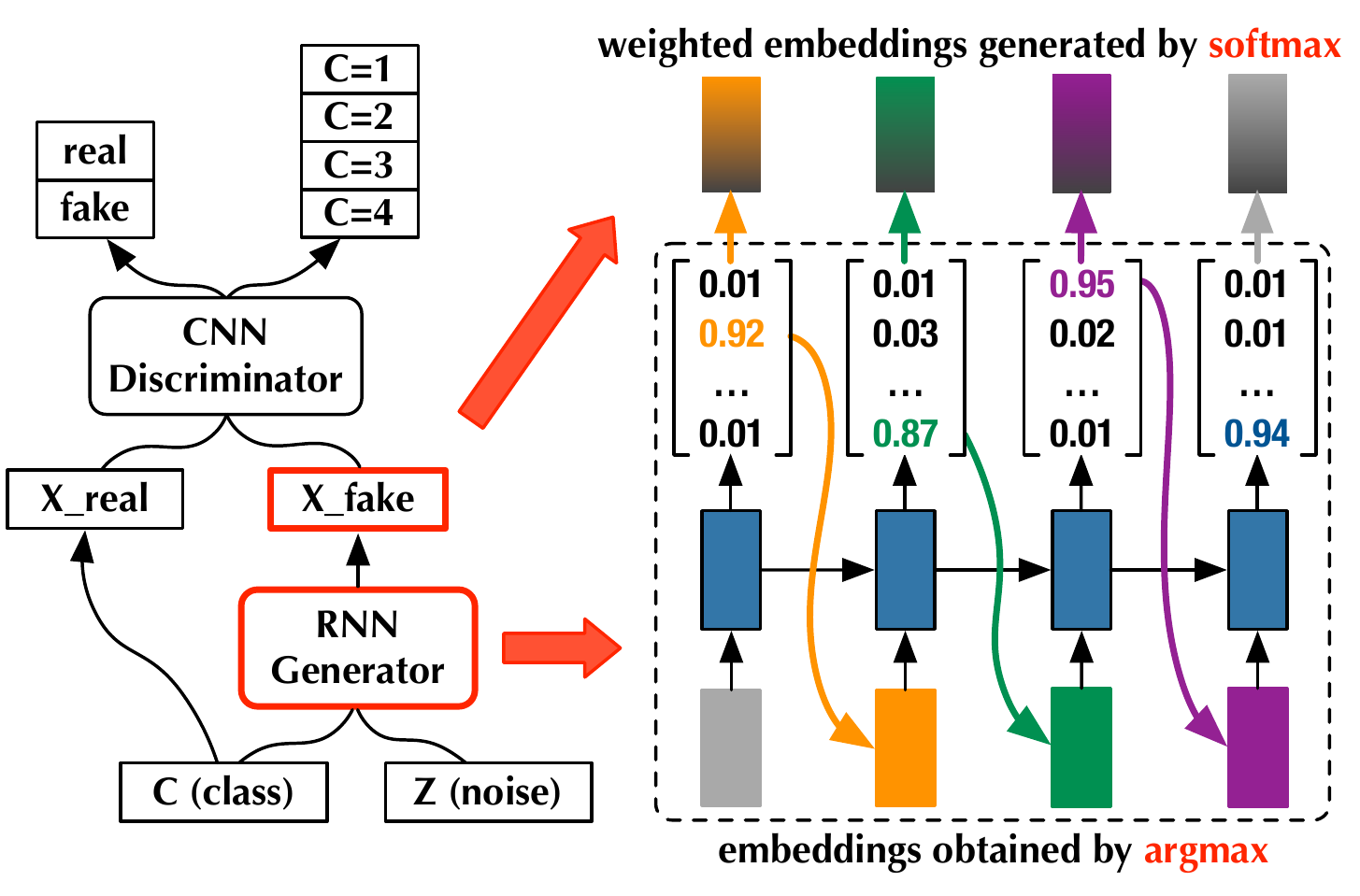}
\caption{System architecture for AC-GAN. (Left) The Generator generates plausible privacy policy sentences ($\mbox{X\_fake}$). 
The Discriminator must learn to differentiate between real and fake sentences as well as predicting the vagueness category ($\mbox{C}$) of the sentences. 
(Right) RNN-generator. A vocabulary distribution is generated for each step. Gumbel-softmax is applied to the distributions to calculate weighted embeddings to be used by the Discriminator (arrows pointing up). 
Argmax is applied to the distributions to retrieve embeddings to be passed to the next step (arrows pointing down).
}
\label{fig:acgan}
\vspace{-0.1in}
\end{figure}

Predicting vague sentences is a nontrivial task due to the complexity and richness of natural language.
We propose to tackle this problem by exploring the auxiliary classifier generative adversarial networks (AC-GAN, Odena et al., 2018\nocite{Odena:2018}).
We choose GAN because of its ability to combine text generation and classification in a unified framework~\cite{Yu:2016,Li:2017:GAN,Gu:2018}. 
Privacy policy sentences are particularly suited for text generation because the policy language is restricted and a text generator can effectively learn the patterns. AC-GAN has a great potential to make use of both human-annotated data and ``fake'' augmented data for classification. 
The system architecture is presented in Figure~\ref{fig:acgan}.
The generator learns to generate ``fake'' privacy policy sentences and sentences exhibiting different levels of vagueness using class conditional probabilities (hence the name auxiliary-classifer GAN). 
The discriminator learns to discriminate among real/fake sentences as well as sentences of different levels of vagueness.
They are jointly trained using a heuristic, non-saturating game loss.
In the following we present the model details.

\subsection{Sentence Generator}

The generator focuses on generating ``fake'' samples that resemble privacy policy sentences of a given vagueness category. 
This is denoted by $P(X|C)$, where $X = \{x_t\}_{t=1}^{T}$ is a sequence of words and $C \in \{1,2,3,4\}$ is a vagueness category.
A vagueness category is randomly sampled in the generation process, and the generator attempts to generate a sentence of that vagueness level.
A typical RNN text generator unrolls the sequence $X$ one word at a time until an end-of-sentence symbol (EOS) is reached.
At time step $t$, it samples a word $x_t$ from a vocabulary-sized vector of probability estimates $P(x_t)$: 
\begin{align*}
& x_t \sim P(x_t) = \mbox{softmax}(\mathbf{a}_t),
\numberthis\label{equ:p_x}\\
& \mathbf{a}_t = \mathbf{W}\mathbf{h}_t + \mathbf{b},
\numberthis\label{equ:a_t}\\
& \mathbf{h}_t = f_{RNN}(\mathbf{h}_{t-1}, x_{t-1}),
\numberthis\label{equ:h_t}
\end{align*}
where $\mathbf{a}_t$ is a vector of activation values and $\mathbf{h}_t$ is the $t$-th RNN hidden state.
We train a neural text generator, implemented as Long Short-Term Memory networks~\cite{Hochreiter:1997}, on a large collection of privacy policy sentences using cross-entropy loss. 
While generating natural language sentences is successfully tackled by recurrent neural networks, the generated sentences are not necessarily vague.
Training the generator only on vague sentences is impractical because there is a limited number of annotated sentences. 
In this paper we introduce a new way of defining class conditional probabilities:
\begin{align*}
x_t \sim P(x_t|C) = \mbox{softmax}(\mathbf{a}_t + \lambda_C \mathbf{v}),
\numberthis\label{equ:x_t_vague}
\end{align*}
where $\mathbf{v}$ is a vocabulary-sized, trainable vector indicating how likely a vocabulary word is vague.
$\lambda_C$ is a coefficient for vagueness category $C$.
The underlying assumption is that a ``clear'' sentence is less likely to contain vague words ($\lambda_C$ is negative), whereas an ``extremely vague'' sentence tends to contain many vague words ($\lambda_C$ is positive).

Finally, the generated ``fake'' sentences, together with ``real'' sentences labelled by human annotators, are fed to the discriminator for training a classifier discriminating between real/fake sentences and sentences of different vagueness levels.
Nevertheless, there remains a critical issue with the current system:
we cannot backpropagate through discrete samples $X$. 
As a result, the generator parameters cannot be properly updated using backpropagation.
To circumvent this issue, we attempt the reparameterization trick with Gumbel-Softmax relaxation~\cite{Gu:2018}.

\vspace{0.05in}
\noindent\textbf{Straight-Through Gumbel-Softmax.}
Two competing issues exist in the RNN generator.
First, the discriminator requires a continuous form for each generated word to keep the entire model differentiable.
Second, the generator requires a discrete choice for each word to generate a sentence, rather than propagating "partial words" through the sequence.
To solve this problem, the softmax distribution of each word is sent to the discriminator, while the argmax over the distribution is sent to the next time step of the generator.
This system is referred to as Straight-Through (ST) Gumbel.

We explain the process of calculating the softmax distribution to send to the discriminator.
To simulate the random-sampling process, the approach applies reparameterization to shift randomness from sampling a discrete variable $x_t$ (Eq.~(\ref{equ:x_t_vague})) to sampling a continuous noise vector $\mathbf{z}_t$ following the Gumbel distribution (Eq.~(\ref{equ:gumbel})). 
The noise vector is added to the activation $\mathbf{a}_t + \lambda_C \mathbf{v}$ to compute the argmax (Eq.~(\ref{equ:x_t_argmax})).
To simulate the argmax operation, a temperature parameter $\tau$ is applied to softmax (Eq.~(\ref{equ:x_t_approx})), where small values of $\tau$ greatly skew the distribution, causing it to peak at the largest value, while still remaining differentiable.
Similar reparameterization is also used for variational auto-encoders~\cite{Kingma:2014}.
\begin{align*}
& \mathbf{z}_t \sim \mbox{Gumbel}(z)
\numberthis\label{equ:gumbel}
\end{align*}
\begin{align*}
& x_t = \mbox{argmax}(\mathbf{a}_t + \mathbf{z}_t + \lambda_C \mathbf{v})
\numberthis\label{equ:x_t_argmax}\\
&P(x_t|C) = \mbox{softmax}(\frac{\mathbf{a}_t + \mathbf{z}_t + \lambda_C \mathbf{v}}{\tau})
\numberthis\label{equ:x_t_approx}
\end{align*}

The generator requires a discrete word to propagate to the next time step of the RNN.
The word with the maximum activation value is chosen as shown in (Eq.~(\ref{equ:x_t_argmax})).
An illustration of ST Gumbel is presented in Figure~\ref{fig:acgan}.

\subsection{Sentence Discriminator}

A sentence discriminator learns to perform two tasks simultaneously.
Given a privacy policy sentence $X$, it predicts a probability distribution over its sources, denoted by $P(S|X)$, where $S$ = $\{\mbox{real, fake}\}$; and a probability distribution over its level of vagueness, denoted by $P(C|X)$, $C$ = \{clear, somewhat clear, vague, extremely vague\}.
The learning objective for the discriminator is to maximize the log-likelihood of making correct predictions on both tasks, denoted by $L_C + L_S$, where $L_C$ and $L_S$ are defined in Eq.~(\ref{equ:L_C}) and (\ref{equ:L_S}).
\begin{align*}
L_C &= \mathop{\mathbb{E}}[\log P(C=c | X_{real+fake})]
\numberthis\label{equ:L_C}\\
L_S &= \mathop{\mathbb{E}}[\log P(S=real | X_{real})] \\
& + \mathop{\mathbb{E}}[\log P(S=fake | X_{fake})]
\numberthis\label{equ:L_S}
\end{align*}

The ground truth vagueness labels $C$ for real sentences are annotated by human annotators.
For fake sentences the labels are randomly sampled in the generation process;
and conditioned on the sampled vagueness labels, fake sentences are generated using $P(x_t|C)$ (Eq.~(\ref{equ:x_t_approx})).
\begin{align*}
L_C' &= \mathop{\mathbb{E}}[\log P(C=c | X_{fake})]
\numberthis\label{equ:L_C_gen}\\
L_S' & = \mathop{\mathbb{E}}[\log P(S=real | X_{fake})]
\numberthis\label{equ:L_S_gen}
\end{align*}

The \emph{generator} is trained to maximize $L_C' + L_S'$ as illustrated in Eq.~(\ref{equ:L_C_gen}-\ref{equ:L_S_gen}).
Intuitively, the generator is rewarded (or punished) only based on the ``fake'' samples it produces.
It is rewarded by generating sentences correctly exhibiting different levels of vagueness, denoted by ($L_C'$).
It is also rewarded by generating sentences that look ``real'' and cannot be easily distinguished by the discriminator ($L_S'$).
Eq.~(\ref{equ:L_S_gen}) corresponds to a heuristic, non-saturating game loss that mitigates gradient saturation~\cite{Goodfellow:2016}.

We experiment with two variants of the discriminator, implemented respectively using the convolutional neural networks (CNN)~\cite{Zhang:2015} and LSTM~\cite{Hochreiter:1997}.
In both cases, the discriminator assigns a source and a vagueness label to each sentence.
The CNN discriminator scans through each sentence using using a sliding window and apply a number of filters to each window.
A max pooling over the sequence is performed to create a feature map for the sentence. 
This feature map is treated as the sentence representation.
It is fed to two separate dense layers with softmax activation to predict $P(C|X)$ and $P(S|X)$ respectively.
In contrast, the LSTM discriminator runs a forward pass through the sentence and uses the last hidden state as the sentence representation.
Similarly, this representation is fed to two dense layers used to predict $P(C|X)$ and $P(S|X)$. 
Both methods produce probability estimations using a shared sentence representation.
Given the scarcity of labelled sentences, this multitask setting is expected to improve the model's generalization capabilities.

\begin{table}[t]
\setlength{\tabcolsep}{9.5pt}
\renewcommand{\arraystretch}{1.1}
\centering
\begin{small}
\begin{tabular}{|l|ccc|}
\hline
& \multicolumn{3}{c|}{\textbf{Word-Level}}\\
\textbf{System} & \textbf{P (\%)} & \textbf{R (\%)} & \textbf{F (\%)} \\
\hline
\hline
Context-Agnostic & 11.30 & \textbf{78.15} & 19.71\\
Context-Aware & \textbf{68.39} & 53.57 & \textbf{60.08}\\
\hline
\end{tabular}
\end{small}
\caption{Results of detecting vague words in privacy policies using context-aware and context-agnostic classifiers.
}
\label{tab:results_word}
\vspace{-0.1in}
\end{table}

\section{Experiments}

We conduct experiments on the annotated corpus using a 5-fold cross validation; 10\% of the training data in each fold are reserved for validation.
In the following sections we present details of experimental settings and report results on detecting vague words and sentences in privacy policy texts.

\subsection{Parameter Settings}
The Xavier scheme~\cite{Glorot:2010} is used for parameter initialization.
For the context-aware word classifier, the bidirectional LSTM has 512 hidden units.
For AC-GAN, the CNN discriminator uses convolutional filters of size $\{3, 4, 5\}$ and 128 filters for each size. 
The LSTM generator and discriminator both have 512 hidden units.
The generator is further pretrained on 82K privacy policy sentences using a 10K vocabulary.
The coefficient $\lambda_C$ is set to $\{-1, 0, 1, 2\}$ respectively for `clear,' `somewhat clear,' `vague,' and `extremely vague' categories.
$\mathbf{v}$ is initialized as a binary vector, where an entry is set to 1 if it is one of the 40 cue words for vagueness~\cite{Bhatia:2016}.
Word embeddings are initialized to their word2vec embeddings and are made trainable during the entire training process.

\subsection{Predicting Vague Words}

We compare {context-aware} with {context-agnostic} classifiers on detecting vague words in privacy policy text.
The goal is to test an important hypothesis: that vagueness is an intrinsic property of words, thus a word being vague has little to do with its context.
Results are presented in Table~\ref{tab:results_word}.

\begin{table}[t]
\setlength{\tabcolsep}{5pt}
\renewcommand{\arraystretch}{1.1}
\centering
\begin{footnotesize}
\begin{fontppl}
\begin{tabular}{|l|p{2.5in}|}
\hline
S1 & ... while we use \textcolor{orange}{\textbf{[reasonable]$_{tp}$}} \textcolor{blue}{\textbf{[efforts]$_{fn}$}} to protect your PII, we can not guarantee its absolute security.\\
\hline
\hline
S2 & We use \textcolor{blue}{\textbf{[third-party]$_{fn}$}} advertising companies to serve \textcolor{orange}{\textbf{[some]$_{tp}$}} of the ads when you visit our web site. \\
\hline
\hline
S3 & The \textcolor{blue}{\textbf{[information]$_{fn}$}} we obtain from \textcolor{blue}{\textbf{[those services]$_{fn}$}} \textcolor{magenta}{\textbf{[often depends]$_{fp}$}} on your settings or their privacy policies, so be sure to check what those are. \\
\hline
\hline
S4 & In the event of an insolvency, bankruptcy or receivership, \textcolor{orange}{\textbf{[personal data may]$_{tp}$}} also be transferred as a business asset. \\
\hline
\end{tabular}
\end{fontppl}
\end{footnotesize}
\caption{Examples of detected vague words in privacy policies. 
$[\cdot]_{tp}$ denotes true positive, $[\cdot]_{fp}$ is false positive, $[\cdot]_{fn}$ is false negative. All unmarked words are true negatives.
}
\label{tab:example_word}
\end{table}

\begin{table}[t]
\setlength{\tabcolsep}{6pt}
\renewcommand{\arraystretch}{1.1}
\centering
\begin{small}
\begin{tabular}{|lr|lr|}
\hline
\multicolumn{2}{|c|}{\textbf{False Alarms}} & \multicolumn{2}{c|}{\textbf{Misses}}\\
\textbf{POS  Tag} & \textbf{Perc. (\%)} & \textbf{POS Tag} & \textbf{Perc. (\%)} \\
\hline
\hline
Adjective & \textbf{37.19} & Noun & \textbf{47.64}\\
Noun & 35.24 & Adjective & 25.07 \\
Verb & 20.53 & Verb & 13.31 \\
Adverb & 4.63 & Adverb & 5.62 \\
Determiner & 1.72 & Determiner & 2.79 \\
\hline
\end{tabular}
\end{small}
\caption{The most frequent part-of-speech (POS) tags appeared in false alarms and misses of detected vague words.
}
\label{tab:stat_pos}
\vspace{-0.1in}
\end{table}

Interestingly, context-agnostic classifier yields a high recall score (78.15\%) despite it ignoring context. 
This result indicates word vagueness can be encoded in distributed word embeddings.
However, the low precision (11.30\%) suggests that context is important for fine-grained analysis. 
While it is possible for experts to create a comprehensive list of vague terms for assessing privacy policies, extra effort is required to verify the tagged vague terms.
Using a context-aware classifier produces more balanced results, improving the F-score from 19.71\% to 60.08\%. 
This indicates that the initial hypothesis is incorrect; rather, context is necessary for detecting vague words.

In Table~\ref{tab:example_word}, we present examples of detected vague words.
The nouns have caught our attention.
The classifier misses several of these, including ``efforts,'' ``information,'' ``services,'' perhaps because there is no clear definition for these terminologies.
In Table~\ref{tab:stat_pos}, we found nouns consist of 47.64\% of all the miss-detected vague words, while adjectives consist of 37.19\% of the false alarms.
There is also an interesting phenomenon.
In S3, ``Information'' and ``those services'' are considered more vague by humans than ``often depends.''
However, if those terms are removed from the sentence, yielding ``The [..] we obtain from [..] often depends on your settings or their privacy policies.'' In this case, the vagueness of ``often depends'' become more prominent and is captured by our system.
It suggests that the degree of vagueness may be relative, depending on if other terms in the sentence are more vague.

\subsection{Predicting Vague Sentences}

In Table~\ref{tab:results_sentence} we present results on classifying privacy policy sentences into four categories: clear, somewhat clear, vague, and extremely vague.
We compare AC-GAN with three baselines: CNN and LSTM trained on human-annotated sentences, and a majority baseline 
that assigns the most frequent label to all test sentences.
We observe that the AC-GAN models (using CNN discriminator) perform strongly, surpassing all baseline approaches.
CNN shows strong performance, yielding an F-score of 50.92\%.
A similar effect has been demonstrated on other sentence classification tasks, where CNN outperforms LSTM and logistic regression classifiers~\cite{Kim:2014,Zhang:2015}.
We report results of AC-GAN using the CNN discriminator.
Comparing ``Full Model'' with ``Vagueness Only,'' we found that allowing the AC-GAN to only discriminate sentences of different levels of vagueness, but not real/fake sentences, yields better results.
We conjecture this is because training GAN models, especially with a multitask learning objective, can be unstable and more effort is required to balance the two objectives ($L_S$ and $L_C$). 
Example sentences generated by AC-GAN are presented in Table~\ref{tab:example_sentence}.

\begin{table}[t]
\setlength{\tabcolsep}{5pt}
\renewcommand{\arraystretch}{1.1}
\centering
\begin{small}
\begin{tabular}{|l|ccc|}
\hline
& \multicolumn{3}{c|}{\textbf{Sentence-Level}}\\
\textbf{System} & \textbf{P (\%)} & \textbf{R (\%)} & \textbf{F (\%)} \\
\hline
\hline
Baseline (Majority) & 25.77 & 50.77 & 34.19\\
LSTM & 47.79 & 50.06 & 47.88\\
CNN & 49.66 & 52.51 & 50.18\\
\hline
\hline
AC-GAN (Full Model) & 51.00 & 53.50 & 50.42\\
AC-GAN (Vagueness Only) & \textbf{52.90} & \textbf{54.64} & \textbf{52.34}\\
\hline
\end{tabular}
\end{small}
\caption{Results on classifying vague sentences.
}
\label{tab:results_sentence}
\end{table}

\begin{table}[t]
\setlength{\tabcolsep}{4.5pt}
\renewcommand{\arraystretch}{1.1}
\centering
\begin{small}
\begin{tabular}{|l|llll|}
\hline
\% (Freq) & {Clear} & {SomeC} & {Vague} & {ExtrV}\\
\hline
\hline
Clear & \textbf{39.4} (477) & 59.8 (723) & 0.7 (8) & 0.2 (2)\\
\hline
SomeC & 12.4 (284) & \textbf{85.2} (1945) & 2.4 (54) & 0.0 (1)\\
\hline
Vague & 3.4 (31) & 89.6 (828) & \textbf{7.0} (65) & 0.0 (0)\\
\hline
ExtrV & 1.2 (1) & 88.9 (72) & 9.9 (8) & \textbf{0.0} (0)\\
\hline
\end{tabular}
\end{small}
\caption{Confusion matrix for sentence classification. The decimal values are the percentage of system-identified sentences that were placed in the specified vagueness class. For example: the item in (row 1, col 2) conveys that 59.8\% of sentences (absolute count is 723) identified by the system as "clear" were actually "somewhat clear" according to humans. 
}
\label{tab:confusion_matrix}
\end{table}

\begin{figure}[t]
\centering
\includegraphics[width=3in]{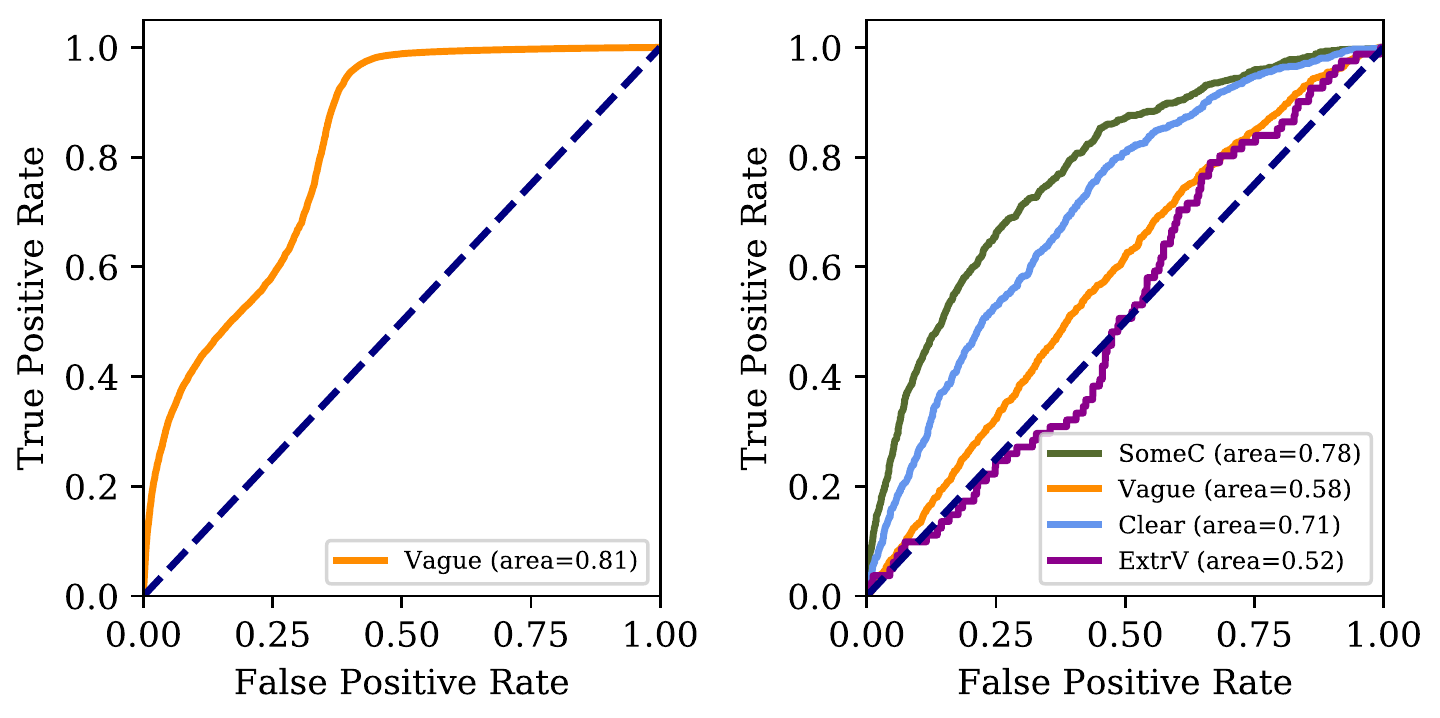}
\caption{ROC curves for classifying vague words (left) and sentences (right).
}
\label{fig:plot_roc}
\vspace{-0.2in}
\end{figure}

Figure~\ref{fig:plot_roc} shows the ROC curves of the four vagueness classes.
Because the dataset is imbalanced, the ROC curves are more informative than F-scores.
The ``clear'' and ``somewhat clear'' classes yield promising AUC scores of 0.71 and 0.78 respectively. 
The ``vague'' and ``extremely vague'' classes are more challenging. 
They are also the minority classes, consisting of 20.5\% and 1.8\% of the annotated data.
Confusion matrix in Table~\ref{tab:confusion_matrix} reveals that the majority of the sentences are tagged as ``somewhat clear,'' while 7.0\% of the vague sentences are tagged as vague.
It suggests more annotated data may be helpful to enable the classifier to distinguish ``vague'' and ``extremely vague'' sentences.
Interestingly, we found there is little correlation between the sentence vagueness score and sentence length (Pearson correlation $r$=0.18, $p<$0.001) while there is a relatively strong correlation ($r$=0.57, $p<$0.001) between sentence vagueness and the number of vague words in it. This finding verifies our hypothesis that vague words seem to increase the perceived sentence vagueness.

\begin{table}[h]
\setlength{\tabcolsep}{4pt}
\renewcommand{\arraystretch}{1.2}
\centering
\begin{footnotesize}
\begin{fontppl}
\begin{tabular}{|l|p{2.4in}|}
\hline
Clear & Our commitment to travian games uses paid services or send an order online.\\
\cline{2-2}
& To learn how important anonymization it, we provide a separate medicare.\\
\hline
\hline
SomeC & Slate use certain cookies and offers.\\
\cline{2-2}
& Visitors who apply us an credit card may sign up.\\
\hline
\hline
Vague & There may take certain incidents various offerings found on various topics; some or all individual has used.\\
\cline{2-2}
& You may modify certain edit or otherwise delete certain features or a similar id will no longer.\\
\hline
\hline
ExtrV & Also, some apps may offer contests, sweepstakes, games or some community where necessary.\\
\cline{2-2}
& If necessary, buying or clarify certain links, certain features of our site may place or some or some features may offer stack or unauthorized access some some functionality.\\
\hline
\end{tabular}
\end{fontppl}
\end{footnotesize}
\caption{Plausible sentences generated by AC-GAN. They exhibit different levels of vagueness. ``SomeC'' and ``ExtrV'' are shorthands for ``somewhat clear'' and ``extremely vague.''
}
\label{tab:example_sentence}
\vspace{-0.1in}
\end{table}

\vspace{0.05in}
\noindent\textbf{Lessons learned.}
We summarize some lessons learned from annotating and detecting vague content in privacy policies, useful for policy regulators, users and website operators. In general, privacy policies are suggested to:
\begin{itemize}[topsep=3pt,itemsep=-1pt,leftmargin=*]
\item provide clear definitions for key concepts. 
Lacking definition is a major source of confusion for the unfamiliar reader. Example concepts include personally identifiable information, personal (non-personal) information, third parties, service providers, subsidiaries, etc.

\item suppress the use of vague words. There are on average 2.5 vague words per sentence in our corpus. The more vague words, the more likely the sentence is perceived as vague ($r$ = 0.57);

\item use sentences with simple syntactic structure to ease understanding. A sophisticated sentence， with vague terms in it, e.g., ``You may request deletion of your personal data by us, but please note that we may be required (by law \emph{or otherwise}) to keep this information and not delete it...'' appears especially confusing to readers.

\end{itemize}

\section{Conclusion}

In this paper we present the first empirical study on automatic detection of vague content in privacy policies.
We create a sizable text corpus including human annotations of vague words and sentences.
We further investigate the feasibility of predicting vague words and sentences using deep neural networks. 
Specifically we investigate context-agnostic and context-aware models for  detecting vague words, and AC-GAN for detecting vague sentences.
Our results suggest that a supervised paradigm for vagueness detection provides a promising avenue for identifying vague content and improving the usability of privacy policies.

\section*{Acknowledgments}
We thank the anonymous reviewers for their valuable comments and  suggestions, which help to improve this paper.
This work is in part supported by an ORC In-House Research Grant awarded to Liu. 

\bibliography{vague,fei,abs_summ,new_ref}
\bibliographystyle{acl_natbib}

\clearpage
\newpage

\end{document}